\documentclass[letterpaper, 10 pt, journal, twoside]{ieeetran}

\IEEEoverridecommandlockouts                              

\usepackage{amsmath} 
\usepackage{amssymb}  
\usepackage[T1]{fontenc}

\usepackage{graphicx}

\DeclareGraphicsExtensions{.png,.pdf,.svg}
\usepackage{caption}
\usepackage{subcaption}
\usepackage{xcolor} 
\usepackage{booktabs}
\usepackage{wrapfig}
\usepackage{gensymb}
\usepackage{ulem}
\usepackage{hyperref}
\usepackage[dvipsnames]{xcolor}

\usepackage{enumitem}

\newcommand{\ours}{VOCALoco}

\title{\ours{}: Viability-Optimized Cost-aware Adaptive Locomotion}

\author{

Stanley Wu$^{1}$, Mohamad H. Danesh$^{1}$, Simon Li$^{2}$, Hanna Yurchyk$^{1}$, Amin Abyaneh$^{2}$, Anas El Houssaini$^{1}$, \\ David Meger$^{1}$, and Hsiu-Chin Lin$^{1,2}$%

\thanks{Manuscript received: June, 15, 2025; Revised October, 9, 2025; Accepted October, 24, 2025.}

\thanks{This paper was recommended for publication by Editor Abderrahmane Kheddar upon evaluation of the Associate Editor and Reviewers' comments.
This work was supported by the Google DeepMind Scholarship and the FRQNT graduate research grants.} 

\thanks{$^{1}$Stanley Wu, Mohamad H. Danesh, Hanna Yurchyk, Anas El Houssaini, David Meger, and Hsiu-Chin Lin are with the School of Computer Science, McGill University, Montreal, Canada
       {\footnotesize \texttt{\textnormal{\{}stanley.wu, mo.danesh, hanna.yurchyk, achraf.elhoussaini\textnormal{\}}@mail.mcgill.ca, 
       \textnormal{\{}david.meger, hsiu-chin.lin\textnormal{\}}@mcgill.ca}}}%

\thanks{$^{2}$Simon Li, Amin Abyaneh, and Hsiu-Chin Lin are with the Department of Electrical and Computer Engineering, McGill University, Montreal, Canada
       {\footnotesize \texttt{\textnormal{\{}xi.yang.li, amin.abyaneh\textnormal{\}}@mail.mcgill.ca}}}%

\thanks{Digital Object Identifier (DOI): see top of this page.}
}%

\newcommand{\rolloutDistance}{d}
\newcommand{\numRollouts}{N}

\newcommand{\R}{\mathbb{R}}
\newcommand{\states}{\mathbf{x}}
\newcommand{\dimStates}{d}
\newcommand{\jointPos}{\mathbf{q}}
\newcommand{\jointVel}{\dot{\jointPos}}
\newcommand{\linVel}{\mathbf{v}}
\newcommand{\angVel}{\boldsymbol{\omega}}
\newcommand{\jointTau}{\boldsymbol{\tau}}

\newcommand{\PolicySet}{\Pi}
\newcommand{\policy}{\pi}
\newcommand{\dimPolicy}{K}
\newcommand{\idxPolicy}{k}

\newcommand{\heightfield}{\mathbf{H}}
\newcommand{\heightfieldLength}{h}
\newcommand{\heightfieldWidth}{w}

\newcommand{\viabilityFun}{\mathcal{V}}
\newcommand{\viability}{v}
\newcommand{\viabilityThreshold}{\epsilon_\viability}

\newcommand{\CoT}{c}
\newcommand{\CoTFun}{\mathcal{C}}

\newcommand{\stanley}[1]{{#1}}

\begin{document}

\IEEEpubid{\begin{minipage}{\textwidth}\centering
\vspace{50pt}
© 2025 IEEE. Personal use of this material is permitted. Permission from IEEE must be obtained for all other uses, in any current or future media, including reprinting/republishing this material for advertising or promotional purposes, creating new collective works, for resale or redistribution to servers or lists, or reuse of any copyrighted component of this work in other works.
\end{minipage}}

\maketitle

\markboth{IEEE Robotics and Automation Letters. Preprint Version. Accepted October, 2025}
{Wu \MakeLowercase{\textit{et al.}}: VOCALoco} 

\begin{abstract}
Recent advancements in legged robot locomotion have facilitated traversal over increasingly complex terrains. Despite this progress, many existing approaches rely on end-to-end deep reinforcement learning (DRL), which poses limitations in terms of safety and interpretability, especially when generalizing to novel terrains. To overcome these challenges, we introduce \ours{}, a modular skill-selection framework that dynamically adapts locomotion strategies based on perceptual input. Given a set of pre-trained locomotion policies, \ours{} evaluates their viability and energy-consumption by predicting both the safety of execution and the anticipated cost of transport over a fixed planning horizon. This joint assessment enables the selection of policies that are both safe and energy-efficient, given the observed local terrain. We evaluate our approach on staircase locomotion tasks, demonstrating its performance in both simulated and real-world scenarios using a quadrupedal robot. Empirical results show that \ours{} achieves improved robustness and safety during stair ascent and descent compared to a conventional end-to-end DRL policy. 

\end{abstract}


\section{INTRODUCTION}
\noindent 

\noindent 
\IEEEPARstart{L}{egged} robots are capable of traversing complex and unstructured terrains that could be inaccessible to wheeled robots \cite{agarwal2023legged, anymal-in-wild}. As a result, the applications of legged robots have been of great interest in problems such as search-and-rescue, remote inspection, and autonomous exploration. While legged robot remains an inherently difficult problem, current research methods have enabled quadruped robots to traverse complex terrains by addressing the challenges of controlling underactuated systems and accurately perceiving an environment that can be highly variable and uncertain. 

Deep reinforcement learning (DRL) has enabled many of the advances in quadruped control over complex terrains \cite{anymal-in-wild}, \cite{anymal-city}, \cite{zhuang2023parkour}, \cite{cheng2023parkour}, \cite{rudin2022advanced}, \cite{anymal-parkour}, \cite{rudin2025parkour}. 
While most DRL methods employ an end-to-end approach, such policies perform poorly outside of their training environment and offer little transparency into their decision-making process. As an alternative, we propose a hierarchical and modular structure that selects multiple specialized policies, each tailored to a specific training distribution. Our method has more control over the behaviour of the robot, as we can tailor locomotion policies to a specific terrain. Our method is also \emph{safer} and \emph{interpretable}, as we evaluate the safety and energy-efficiency of each policy, allowing us to assess the behaviour of the robot at runtime. However, deploying multiple low-level policies requires a robust high-level strategy to switch between them. 



Inspired by animal locomotion, where gait transitions occur naturally in response to environmental demands, prior work examined adaptive gait selection driven by base velocity. \cite{yang2022fast}, \cite{fu2021minimizing}, \cite{liang2024adaptive}, \cite{bellegarda2022cpg}, \cite{bellegarda2024allgaits}, \cite{Shafiee2024}. While these methods are effective on flat and slightly uneven terrain, we focus on switching between gaits or skills on unstructured terrain. Few works have studied \textit{non} end-to-end skill switching on unstructured terrain, namely \cite{Shafiee2024}, \cite{anymal-parkour}. We improve on \cite{anymal-parkour} by simplifying the high-level skill selection with a supervised learning approach inspired by \cite{Shafiee2024}.




\begin{figure}[t]
    \centering
    \includegraphics[width=\columnwidth]{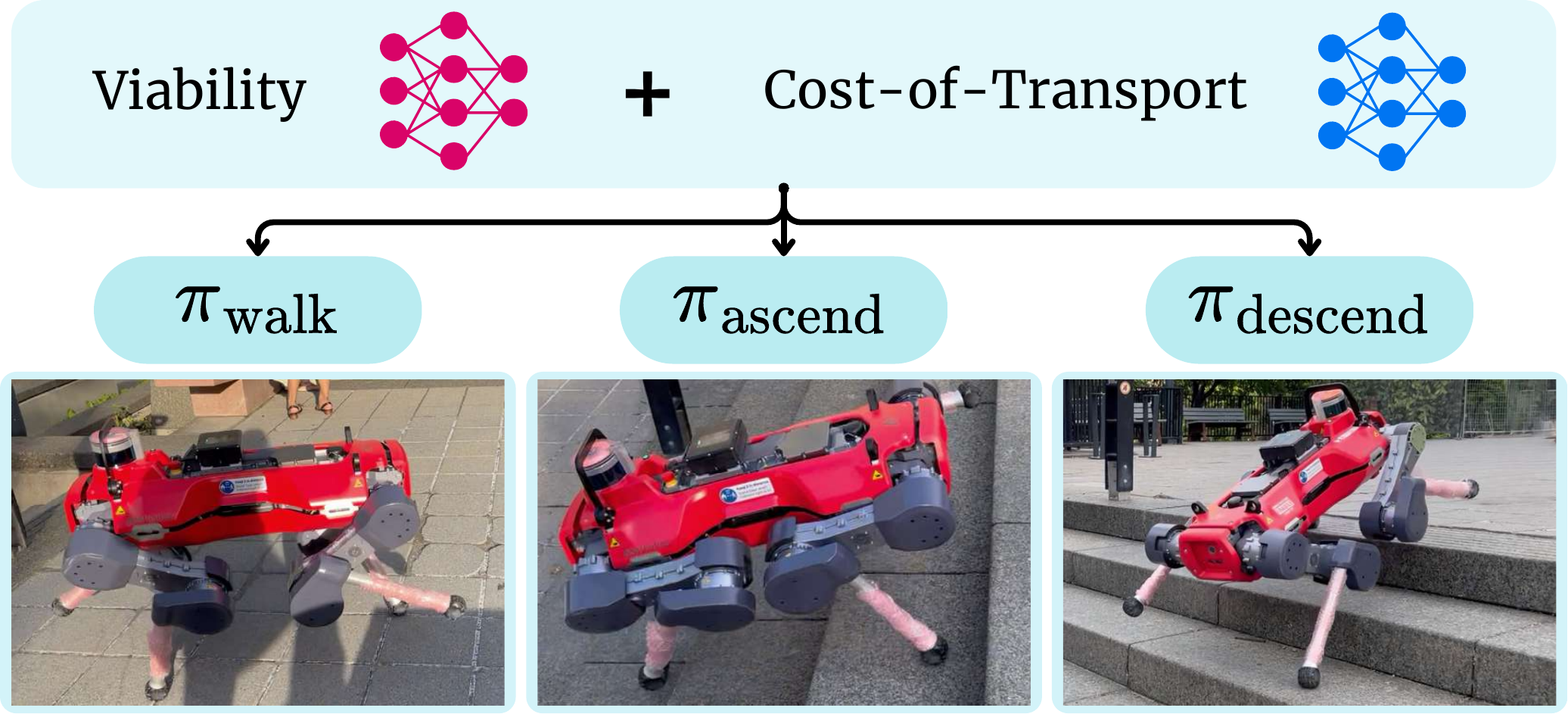}
    \caption{Overview of \ours{}. Given a heightmap of the local terrain, two high-level modules predict: (i) the viability and (ii) the Cost-of-Transport (CoT) of executing each skill of executing each skill. With both predictions at hand, we first filter unsafe skills. Then, among the safe skills, we select the skill with the lowest predicted energy expenditure as the final policy to execute on the robot. The three example images show the ANYmal-D robot switching between different policies depending on the terrain type.}
    \label{fig:intro}
\end{figure}

We propose \textbf{V}iability-\textbf{O}ptimized \textbf{C}ost-aware \textbf{A}daptive \textbf{Loco}motion (\ours{}), a hierarchical skill selection framework that uses the robot's perception of its local terrain to determine the most suitable locomotion policy. Assuming access to a set of low-level locomotion skills, our approach employs a high-level decision-maker to predict the viability and energy cost of each skill on the local terrain, enabling adaptive skill selection, conceptually illustrated in Figure~\ref{fig:intro}.
Our key contributions are as follows:
\begin{itemize}[topsep=0pt]
    \item We introduce a method to predict the viability and energy cost of executing different locomotion policies on perceived terrain structures, enabling safe and energy-aware skill switching.
    \item Our modular framework provides greater {\em interpretability} than standard end-to-end policies, allowing experts to fine-tune or extend the set of low-level skills.
    \item All components are trained entirely in simulation using automatically generated labels, requiring no manual annotation or human intervention.
    \item We validate our approach on the ANYmal-D \cite{anymal-d} robot in both simulation and the real-world with a zero-shot deployment.
\end{itemize}

\section{Related Work}

\subsection{DRL for Legged Robots}
\noindent 
In recent years, DRL has shown impressive results for legged robots. Part of the success can be attributed to closing the sim-2-real gap with domain and dynamics randomization \cite{peng2018sim}, \cite{bipedal-blind-stair}, learning to estimate terrain and environment properties \cite{chen2020learning}, \cite{lee2020learning}, \cite{kumar2021rma}, learning with GPU-acceleration and a game-style curriculum \cite{legged_gym}, and simulating realistic actuator dynamics \cite{actuator-net}. 

Several works have demonstrated that incorporating environmental perception for DRL locomotion is essential for avoiding obstacles, traversing challenging terrain and generalizing to novel terrain \cite{agarwal2023legged}, \cite{yang2023neural}, \cite{yang2022learning}, \cite{visual-cpg-rl}, \cite{risky}, \cite{control-and-state}, \cite{rloc}, \cite{dreamwaq}, \cite{dreamwaq++}. The benefits of perception to generalize to diverse terrains has also been demonstrated in long-distance locomotion \cite{anymal-in-wild}, \cite{anymal-city}, in parkour terrains \cite{zhuang2023parkour}, \cite{cheng2023parkour}, \cite{rudin2022advanced}, \cite{anymal-parkour}, \cite{rudin2025parkour}, and in other highly unstructured terrains \cite{risky}, \cite{crawl}, \cite{kareer2022vinl}. 
However, these approaches typically rely on a single end-to-end policy, which limits their flexibility: they are difficult to interpret and hard to adapt to new tasks.

One work on quadruped locomotion for parkour \cite{anymal-parkour} employed a similar hierarchical architecture, where the high-level policy selects low-level locomotion policies. However, their high-level policy is a deep reinforcement learning policy, lacking interpretability. In contrast, our high-level policy is composed of two modules predicting the viability and energy consumption of each low-level policy, resulting in a more explainable and safety-aware approach. Furthermore, we used supervised learning to train our high-level modules, which avoids the complexity of using DRL.

\subsection{Geometric Traversability}
\label{geo-trav}
\noindent One popular approach to estimate traversability on rigid terrain structures is by rolling out robot trajectories in simulation and automatically collecting synthetic traversability data \cite{ground_traversability}, \cite{yang2021real}, \cite{voxel-isaac-traversability}, \cite{muhamad2024robust}. The advantage of this approach is the ability to generate infinite data automatically without any human labelling. We use this approach to estimate geometric traversability for each of our policies. Specifically, we collect simulation data similarly to \cite{voxel-isaac-traversability} and extend this approach with skill selection.

\subsection{Skill and Gait Switching for Legged Robots}
\label{gait-switch}
\noindent
Recent studies explored quadruped skill and gait switching. \cite{chamorro2024reinforcement} and \cite{margolis2023walk} both trained DRL policies that are capable of adapting their behaviours to traverse different terrain types. However, both frameworks required a human operator to vary some parameters of their methods at runtime.

Other prior works have explicitly embedded traversability estimation into gait switching \cite{zenker2013visual}, \cite{elnoor2024pronav}. However, these studies primarily focused on determining whether a terrain is traversable for navigation purposes. 
Further works have implicitly adapted locomotion behaviour based on terrain structure and difficulty, by adapting the footsteps~\cite{belter2019single} or adapting the whole-body movement~\cite{jenelten2022tamols}.

Drawing parallel analogies to how and why animals switch between different gaits, studies have demonstrated the influence of robot energy consumption, or CoT, on gait patterns \cite{yang2022fast}, \cite{fu2021minimizing}, \cite{liang2024adaptive}. The link between CoT and gait switching has been further highlighted by a series of work on central-pattern-generators (CPG) for quadruped robots \cite{bellegarda2022cpg}, \cite{bellegarda2024allgaits}, \cite{Shafiee2024}. However, these works primarily focus on gait switching based on base velocities on flat or slightly uneven terrain, instead of more complex terrains. \cite{Shafiee2024} address this issue by investigating biomechanical factors inducing gait switching on terrains with consecutive gaps. In contrast, \ours{} learns to predict these biomechanical factors explicitly, using viability and CoT metrics estimated from perception, to perform skill selection on more diverse terrain structures.



\section{Problem Statement and Definitions}
\label{sec:preliminaries}

\subsection{Problem Formulation}
\noindent We consider a legged robot whose full state is denoted by $\states \in \mathbb{R}^{\dimStates}$. This state vector comprises the robot's joint positions $\jointPos$, joint velocities $\jointVel$, joint torques $\tau$, base linear velocity $\linVel$, base angular velocity $\angVel$, and a local terrain representation $\heightfield \in \mathbb{R}^{\heightfieldLength \times \heightfieldWidth}$.


\stanley{Our local terrain representation $\heightfield$ is a {\em heightfield}. To ensure it covers the local terrain around and under the hind legs of the robot to prevent it from getting stuck on certain terrain types, we choose a rectangular region that expands $2m$ forward and $1m$ backward from the robot's base, and a width of $1m$. The points in the heightfield are spaced by $10cm$, yielding a $31 \times 11$ matrix.}

The robot is equipped with a repertoire of $\dimPolicy$ distinct locomotion skills, represented as a set of parameterized policies $\PolicySet = \{\policy_1, \policy_2, \dots, \policy_\dimPolicy\}$. These policies may be obtained via diverse learning or optimization strategies, including DRL, imitation learning, or trajectory optimization. We impose no restrictions on the source or training method of these policies.

Given the current state $\states$ and observation $\heightfield$, our objective is to design a \emph{skill selection mechanism} that chooses the most energy-efficient policy $\policy_\idxPolicy \in \PolicySet$ from among those deemed viable i.e., capable of safely and stably executing locomotion in the current terrain context.

\begin{figure*}[t!]
    \centering
    \includegraphics[width=\textwidth, keepaspectratio]{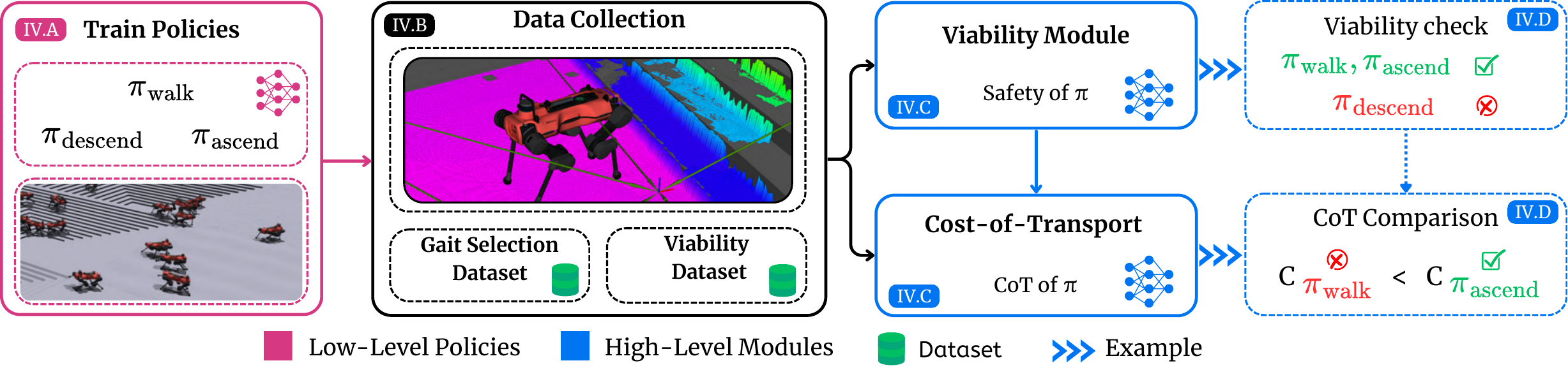}
    \caption{In \ours{}, we start by training low-level locomotion policies: a walking policy, an ascending policy, and a descending policy. Then, we perform rollouts with these policies, collecting data that will train the high-level policies: the viability and the CoT modules.}
    \label{fig:method-overview}
\end{figure*}

\subsection{Viability}
\label{sec:background-viability}
\noindent
The term \emph{viability} refers to the ability of the robot to achieve its designated task. In \ours{}, we define the notion of viability as the likelihood that a locomotion skill achieves our desired task of moving forward over a fixed horizon safely without any base collisions. 

Several related works discussed in Section~\ref{geo-trav} adopt the term \emph{traversability} to describe terrain assessment in order to avoid dangerous regions. In this work, we deliberately use the term viability instead. This choice reflects the objective of \ours{} to focus on estimating whether the robot can move forward over a fixed horizon without a fatal collision. Furthermore, this terminology is consistent with the usage in \cite{Shafiee2024}, which similarly emphasizes progress-oriented viability rather than a comprehensive traversability analysis.

\subsection{Cost of Transport}
\label{sec:background-cot}
\noindent
CoT measures the energy required to move a unit weight over a unit distance. In the context of gait switching, CoT plays a central role in determining when and why a robot should switch gaits.
Formally, the CoT is defined as
\begin{equation}
    \CoT=\frac{\sum_{t=0}^T \jointTau_t^{\top} \jointVel_t \Delta t}{mg \rolloutDistance}
    \label{equ:cot}
\end{equation}
where $T$ is the total timesteps required to complete the task, $\jointTau_t$ is total joint torques computed at time $t$, $\jointVel_t$ is the joint velocities at time $t$, and $\Delta t$ is the simulation timestep, $m$ is the mass of the robot, $g$ is the gravity constant, and $\rolloutDistance$ is the distance travelled during the period $T$.

Prior work has demonstrated that different gaits (e.g., walking, trotting, galloping) have different CoT profiles depending on the terrain, speed, and payload (section \ref{gait-switch}). An underlying assumption is that animals can reduce their overall energy consumption by switching to the gait with the lowest CoT for the current conditions~\cite{farris2012mechanics}, \cite{summerside2018contributions}, \cite{srinivasan2006computer}. In our work, we assume that locomotion policies, whether obtained through reinforcement learning or optimal control, are energy efficient on the terrain they were trained. We will learn a model that predicts the CoT over a short fixed horizon. 


\section{METHODOLOGY}\label{sec:method}

\noindent 
\ours{} follows a hierarchical structure to safe and robust quadruped locomotion, as outlined in Figure~\ref{fig:method-overview}. 
Assuming the robot is equipped with different low-level locomotion skills, we proposed a high-level decision-making component that reads the robot's perception of the local terrain to evaluate each available locomotion skill. Over a short horizon, the viability module estimates the likelihood of successful traversal (i.e., viability), while the CoT module predicts the expected energy cost. Based on these predictions, we select the skill that is both safe and energy-efficient. We first train a set of low-level locomotion policies specialized for different terrain types in Section~\ref{sec:low_level_policies}. These policies are then rolled out in simulation to generate synthetic data in Section~\ref{sec:data-collection}, and to ultimately train the high-level viability and CoT modules in Section~\ref{sec:viability_cot_learning}.

\subsection{Low-Level Policies}
\label{sec:low_level_policies}
\noindent While our method makes no assumption about how the low-level policies are gathered, we demonstrated our work with policies $\policy_\idxPolicy$ trained with DRL in the Legged Gym simulator \cite{legged_gym}, which enables efficient training via multi-environment parallelization. For policy optimization, we adopt Proximal Policy Optimization (PPO)~\cite{schulman2017proximalpolicyoptimizationalgorithms}, a widely used DRL algorithm. We focus on learning policies for three distinct locomotion skills: (1) walking on flat terrain, (2) ascending stairs, and (3) descending stairs. Further details of \ours{}'s training setup are delineated in Section~\ref{sec:exp}.







\subsection{Data Collection}
\label{sec:data-collection}

\begin{figure*}[t]
    \centering
    \includegraphics[width=\textwidth, keepaspectratio]{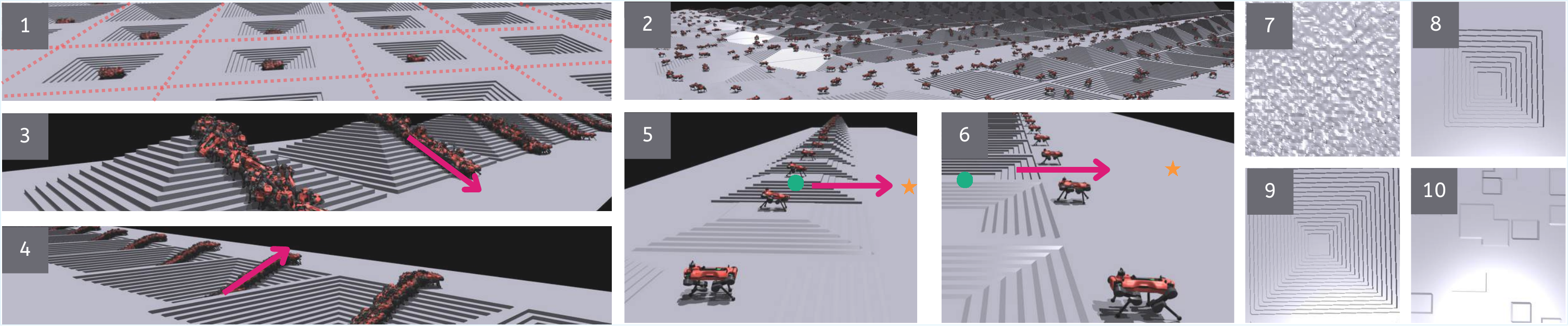}
    \caption{Simulation terrains. 
    (1) An example of a low-level locomotion policy training environment. The red dots denote the boundaries of the terrain cell that each robot cannot cross. 
    (2) Policy rollouts on different terrain types to gather data for the high-level modules. 
    (3) \stanley{Descending} staircase environment.
    (4) \stanley{Ascending} staircase environment.
    (5, 6) Evaluation environments. The terrains with transitions from flat to stairs and vice versa. The green dot represents the spawn position of the robot, and the star represents the target. 
    (7) Rough terrain. 
    (8) Stairs up.
    (9) Stairs down.
    (10) Discrete terrain.}
    \label{fig:sim}
\end{figure*}


\noindent 
The data is collected by running various scenarios in simulation using Isaac Gym \cite{isaac-gym}, to build a dataset of heightfields paired with their corresponding geometric viability and CoT for each low-level policy. We generate environments across three terrain types: (1) flat and uneven terrain, (2) ascending staircases, and (3) descending staircases. Figure~\hyperref[fig:sim]{\ref*{fig:sim}.2}, \hyperref[fig:sim]{\ref*{fig:sim}.3}, and \hyperref[fig:sim]{\ref*{fig:sim}.4} show an example of the terrains we used to collect data. 
For each terrain type, we vary the {\em difficulty level} to generate extra environments (e.g., obstacle size, step height, etc). 

We randomly spawn the robot in simulation and align the robot's roll and pitch with the terrain underneath it. Since the terrain and the initial pose of the robot are random, some combinations of initial positions may be infeasible to stand on (e.g., starting the robot at the edge of the staircase). 
We verify that the spawn position is valid by waiting for 1 second and checking if the robot's base is close to the desired spawn position, the robot's yaw is aligned with the desired yaw, and no base collisions are detected.
Otherwise, another initial pose is generated.

\subsubsection{Collecting Viability Data}
If the robot has a valid initial pose, we take the local heightfield of the environment.
For each valid initial pose, we rollout each policy on the same initial pose by moving the robot over a fixed horizon distance using a fixed velocity of $0.6m/s$. We choose $0.6m/s$ because on the real ANYmal-D, using a faster velocity could potentially lead to more noisy heightfields \cite{legged_gym}.
If the robot successfully reaches the target and no base collisions are detected, we label that data point as a \emph{success}.
If the robot did not reach the target \stanley{with a time limit of 4 seconds}, which could happen due to a collision or a terrain structure that steered the robot off its path, we label the data point as a \emph{failure}. 
The same procedure is repeated for $\numRollouts$ rollouts to obtain the success rate ($\frac{\text{number of successes }}{\numRollouts}$). 

For each sample, each value in the heightfield in simulation is injected with noise coming from a uniform distribution from -10cm to 10cm, which are the default noise values from \cite{legged_gym}. Furthermore, we normalize the heightfield by subtracting the height value at its center from all other values. Specifically, if the center value is $x$, and a given heightfield value is $h$, we compute the normalized value as $ h' = h - x $. This centers the heightfield around the robot's local elevation.

\subsubsection{Collecting CoT Data}
We follow a similar procedure to collect CoT data, where we rollout the policy and calculate its CoT using Equation~\ref{equ:cot}. However, unlike in the viability data collection case, we set $\numRollouts = 1$. Specifically, for a sampled heightfield, we do 1 rollout only since we are not measuring the success rate. In addition, the measured CoT values can be noisy. When a robot switches from a spawn state to a rollout state, it takes a maximum of $0.5s$ for the robot to accelerate to $0.6m/s$, depending on the low-level policy running on the robot, the spawn pose, and the spawn location. Therefore, to reduce noise in the data, we start measuring the CoT after the first $0.5s$ of the rollout. The rollout distance remains at $1.5m$. Any rollout that results in a base crash is marked invalid and excluded from the dataset.

\subsubsection{Horizon distance}
While shorter rollout distances could better capture proximal terrain features to the robot, we set the rollout distance to $1.5m$ to avoid excessively frequent policy switching that can destabilize skill selection. Empirically, $1.5m$ also produces viability and CoT measurements with acceptable noise levels.

\begin{figure}[t]
    \centering
    \includegraphics[width=0.67\columnwidth, keepaspectratio]{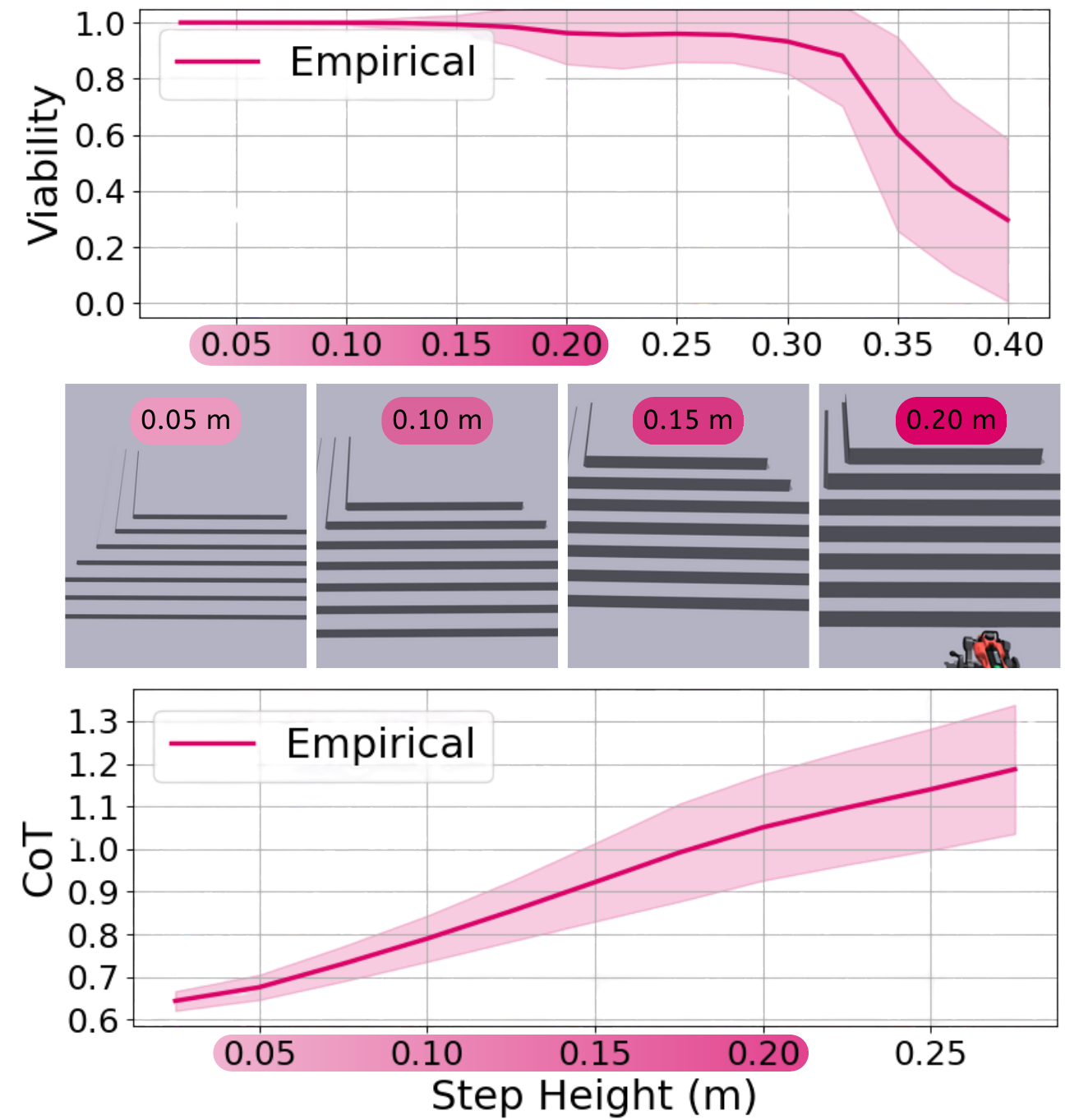}
    \caption{Visualization of (middle) simulated environments with various difficulty levels and the collected (top) viability and (bottom) CoT data from rolling out the descent policy across different difficulty levels. The shaded region represents the standard deviation.}
    \label{fig:sim-and-data}
\end{figure}

Figure~\ref{fig:sim-and-data} illustrates a representative example of the collected data.
Terrains with different difficulty levels (step height) are generated (middle).
The viability (top) and CoT (bottom) are collected by rolling out a policy in these terrains.
As terrain difficulty increases, viability decreases while the cost of transport increases, which are both expected.
Note that the plot for CoT stops at step height = 0.275 as the policy is no longer viable beyond this point.

\subsection{Learning Viability and Cost-of-Transport}
\label{sec:viability_cot_learning}
\noindent
Given the data collected from Section~\ref{sec:data-collection}, 
we aim to learn a mapping that predicts viability and CoT given a heightfield.
We take the policy viability as a regression problem on heightfield data. 
For each policy $\pi_\idxPolicy$ in $\Pi$, we aim to learn a predictive function that estimates the viability $\viabilityFun^\idxPolicy(\heightfield)$. 

Each $\viabilityFun^\idxPolicy(\mathbf{H})$ takes a heightfield $\mathbf{H}\in \mathbb{R}^{\heightfieldWidth \times \heightfieldLength}$ and regresses to a value between 0 and 1, where the higher the estimate, the more likely that traversing the local terrain of the robot is viable under a policy $\pi_\idxPolicy$: $\viabilityFun^{\idxPolicy}(\heightfield): \mathbb{R}^{\heightfieldWidth \times \heightfieldLength} \rightarrow [0, 1]$.
For each policy, we also learn a predictive function that estimates the CoT: $\CoTFun^{\idxPolicy}(\heightfield): \mathbb{R}^{\heightfieldWidth \times \heightfieldLength} \rightarrow \R^+$.


\subsection{Skill Selection}
\noindent
Our high-level component selects the optimal low-level policy given the robot’s local terrain. It takes the heightfield $\heightfield$ as input and predicts, for each policy $\policy_k$, its viability $\viabilityFun^k(\heightfield)$ and CoT $\CoTFun^k(\heightfield)$. Policies with a viability lower than a predefined threshold $\viabilityFun^k(\heightfield) < \viabilityThreshold$ are discarded as unsafe. Among the remaining candidates, the module would select the policy with the lowest predicted CoT. \stanley{Unlike prior works using CoT implicitly as a low-level reward term during training \cite{legged_gym} \cite{dreamwaq} \cite{dreamwaq++} \cite{fu2021minimizing} \cite{liang2024adaptive}, VOCALoco predicts CoT explicitly at runtime to compare multiple pretrained skills and select the most energy-efficient one.}

\section{EXPERIMENTS}

\newcommand{\flatPolicy}{\pi_{\text{flat}}}
\newcommand{\ascendPolicy}{\pi_{\text{ascend}}}
\newcommand{\descendPolicy}{\pi_{\text{descend}}}


\subsection{Experiment Setup}\label{sec:exp}
\noindent We begin with three locomotion tasks, walking, ascending stairs, and descending stairs. For this, 3 low-level policies are trained, together with their corresponding viability and CoT prediction model, resulting in 6 CNN in total.

\label{legged-gym-policies}

\subsubsection{Training Terrain-Specialized Locomotion Policies}

\noindent
\stanley{As discussed in Section~\ref{sec:method}, three terrain types are used for policy learning and data collection. The walking policy trains on $60\%$ flat, $20\%$ discrete, and $20\%$ rough terrains. The climbing-up and climbing-down policies each use $75\%$ stair terrains (ascending or descending, respectively) and $25\%$ rough/discrete terrains. Mixing terrain types per policy yields more natural gaits.}



\stanley{The rest of the environment settings are inherited from \cite{legged_gym}, including the curriculum learning, noise in the observation space, the action space, and the domain randomization ranges.}.
We also use similar reward terms as \cite{legged_gym} to train policies that are highly specialized in their respective terrain\footnote{\stanley{Details are available at: \url{https://sites.google.com/view/vocaloco}.}}.

\subsubsection{Heightfield}
The robot processes depth camera images into elevation maps ~\cite{fankhauser2014robot}, from which we query to produce heightfields. One issue is that occlusions in the heightfield could potentially cause incorrect predictions. This issue becomes especially problematic when the robot is ascending stairs and has no knowledge of the terrain structure beyond the top of the stairs, potentially causing the viability CNNs to return a conservative estimate. To address this, we perform a forward fill along each column of the heightfield, replacing occluded cells (NaN) with the last valid value observed behind them. This operation proceeds from the back of the robot toward the front, allowing us to propagate known height values forward into occluded regions.

\begin{figure}[t]
    \centering
    \includegraphics[width=0.92\columnwidth]{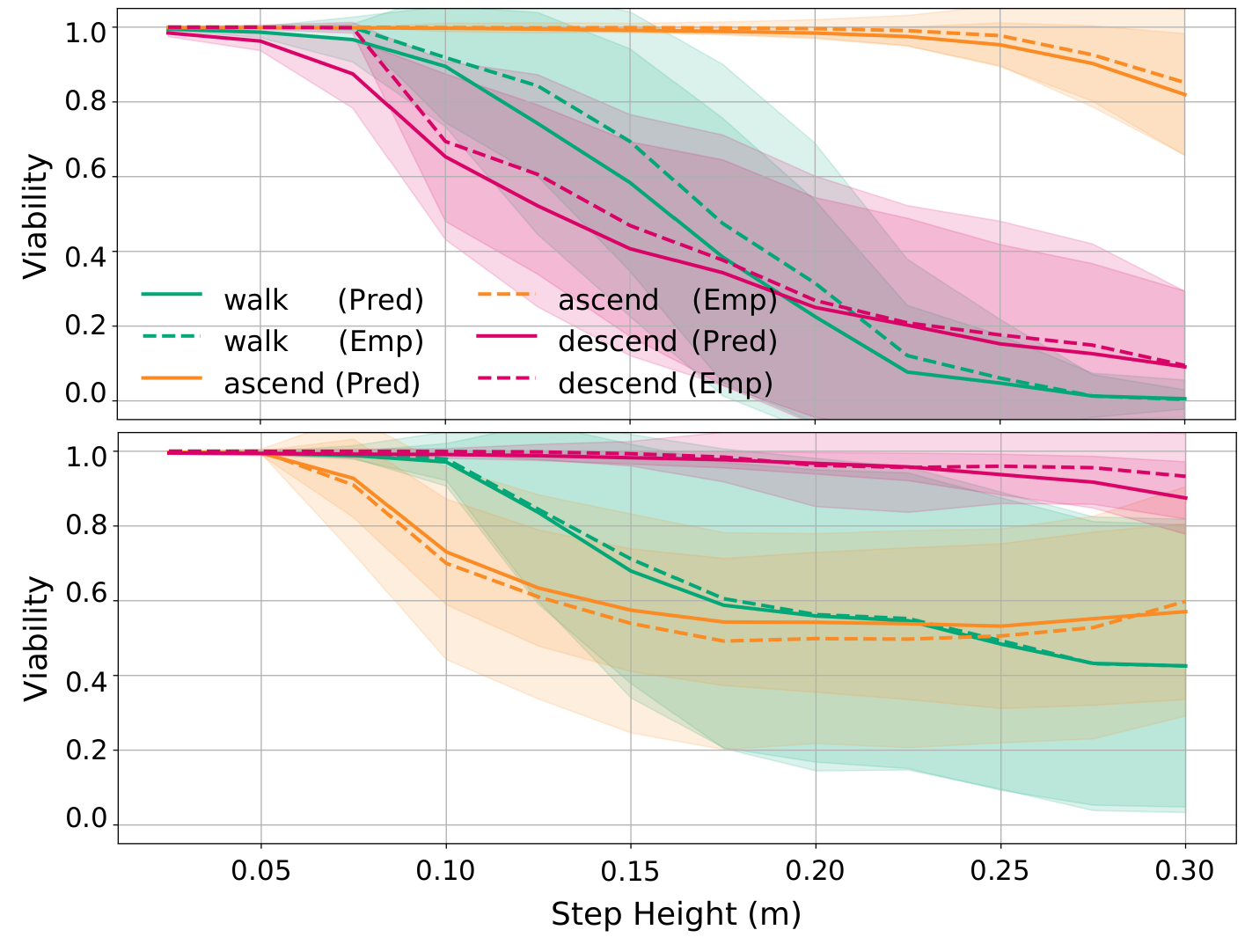}
    \caption{The collected and the predicted viability across terrains with different step heights for (Top) ascending staircase environment (Figure~\hyperref[fig:sim]{\ref*{fig:sim}.3}) and (Bottom) descending staircase environment (Figure~\hyperref[fig:sim]{\ref*{fig:sim}.4}). 
    }
    \label{fig:viability-emp-vs-pred}
\end{figure}

\subsubsection{Implementation Details}
Each CNN in both the viability and CoT modules comprises 3 convolutional layers with 4, 8, and 8 channels, respectively. All convolutional layers use a 3$\times$3 kernel, stride 1, and padding 1. A MaxPool layer (kernel 2) downsamples between the second and third convolutional layers. After these 3 convolutions, the feature map is passed through 2 fully connected layers of 128 neurons each before reaching a single output head. ReLU activations are applied to all hidden layers, but only the Viability CNN applies a sigmoid at the last logit. During training, we use mean-squared error loss and stochastic gradient descent with a learning rate of 0.005 and momentum of 0.9. The dataset for each CNN has 100k samples. 

\stanley{
Our experiments were carried out on a NVIDIA 4090 GPU with VRAM of 24GB.
In our case, collecting 100K viability and CoT samples takes around 4 hours and 12 minutes, respectively. Training the CNNs with 100K samples also takes around 12 minutes. We can parallelize the tasks, taking a total of 8-10 hours for data collection and training.}


We set the viability thresholds to ${\viabilityThreshold}_{\text{, walk}} = 0.95$, ${\viabilityThreshold}_{\text{, ascend}} = 0.925$, and ${\viabilityThreshold}_{\text{, descend}} = 0.925$, both in simulation and on the real robot. Selecting these values trades off risk-averse versus risk-seeking behavior in our viability CNN: higher thresholds yield more conservative behavior, leading to more false negatives (safe terrain flagged as unviable), whereas lower thresholds produce more aggressive behavior, resulting in more false positives (unsafe terrain flagged as viable). The chosen values strike a balance between avoiding unnecessary stops and preventing collisions or stalls.

We run the high-level modules at 50 Hz and we pass each predicted skill into a sliding window of length 10. When all skills in the window agree, we execute that skill on the robot. 
This filtering smooths transitions: shorter windows lead to redundant switching, while longer windows introduce lag.
Empirically, we found that using a window size of 10 enables timely policy transitions while avoiding redundant toggling.



\subsection{High-Level Module Performances}
\noindent 
We start our analysis by validating the performance of the learned viability modules. 
Figure~\ref{fig:viability-emp-vs-pred} shows the collected data and the predicted viability across terrains of increasing difficulty on ascending  (Figure~\hyperref[fig:sim]{\ref*{fig:sim}.3}) and descending staircase environments  (Figure~\hyperref[fig:sim]{\ref*{fig:sim}.4}). 
For each step height (incremented by $2.5cm$), we collect a test set of 500 samples.

For all policies, we observe that as the step heights increase, the viability decreases. While this result is expected, it is crucial to verify that our viability CNNs are well-calibrated on the terrain structures of interest. Accurate calibration allows us to set custom thresholds $\viabilityThreshold$ for each policy, thereby reliably determining whether a terrain segment is safe or unsafe to traverse.


\begin{figure}[t]
    \centering
    \includegraphics[width=0.88\columnwidth]{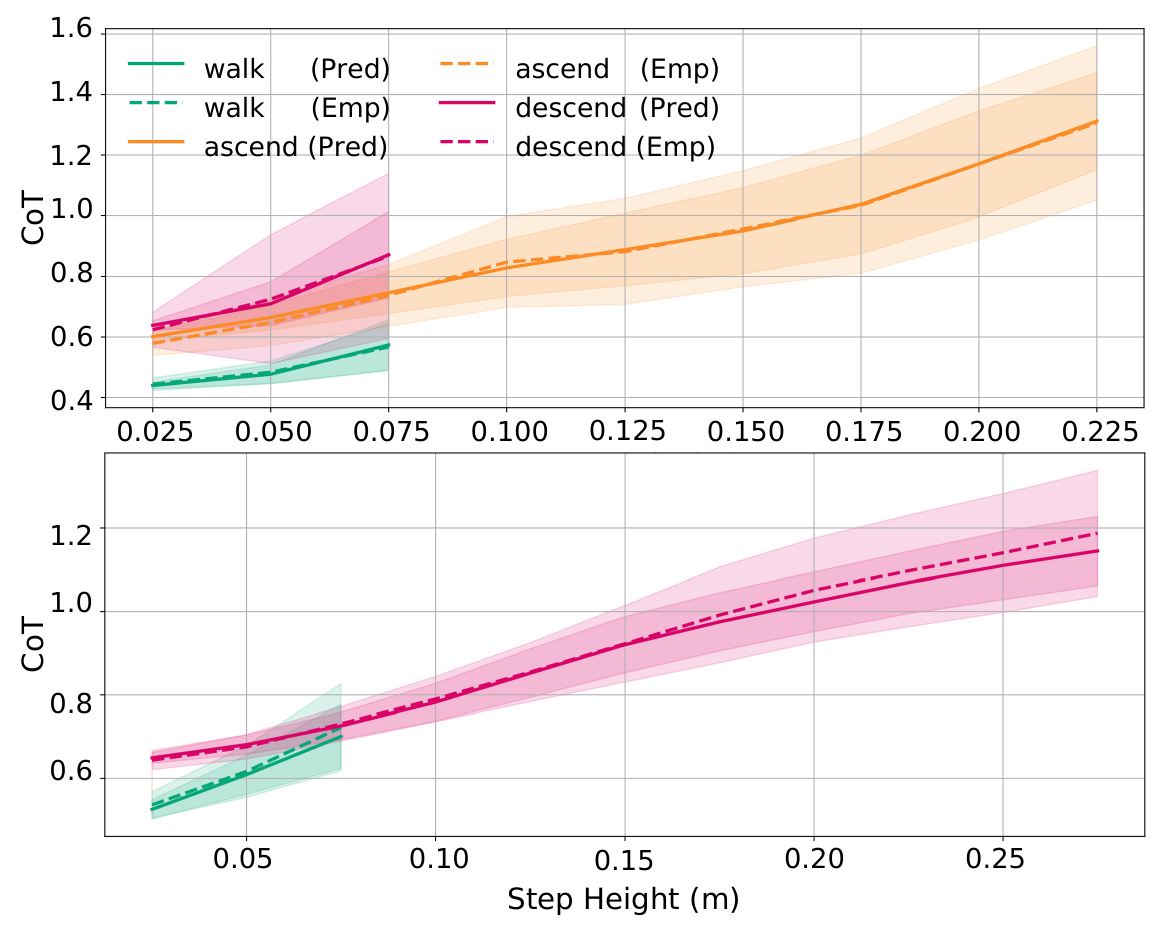}
    \caption{The collected and predicted CoT across terrains with different step heights for (Top) ascending staircase environment (Figure~\hyperref[fig:sim]{\ref*{fig:sim}.3}) and (Bottom) descending staircase environment (Figure~\hyperref[fig:sim]{\ref*{fig:sim}.4}). We only plot data points if the policy is viable during rollouts (with threshold = 90\%).}
    \label{fig:learned-cot}
\end{figure}

Similarly, Figure~\ref{fig:learned-cot} shows the predicted CoT and the variance over 500 testing data. First, we observe the expected general trend of increasing CoT as the step heights increase. 
We further note that the walking policy incurs a lower CoT than the ascend and descend policies on very low step heights. 
This observation is consistent with expectations, as the walking policy generally generates lower swing trajectories, resulting in reduced energy expenditure relative to the other policies.
Note that, we only plot data points if the policy is (up to 90\%) viable during rollouts. 
The outcome of the ascend policy on the descending staircase environment is omitted, because every test set at every height contained more than 10\% of crashes.

Note that, despite fixing the step height for each test set, the data will still be diverse since the initial position and orientation of the base are randomly generated, and each test may involve terrain transitions from flat to stairs and/or from stairs to flat terrain. 
Therefore, there is a variance across most subplots in Figures \ref{fig:viability-emp-vs-pred} and \ref{fig:learned-cot}. 
The variance of the test set and the predictions also increases as the step heights increase. This is also expected, since more difficult terrain leads to greater low-level policy instability and overall unpredictability in the robot's behaviour, especially when operating in an environment outside its training distribution.

\subsection{Performance in Simulation} 

\begin{figure}[t]
    \centering
    \includegraphics[width=0.92\columnwidth, keepaspectratio]{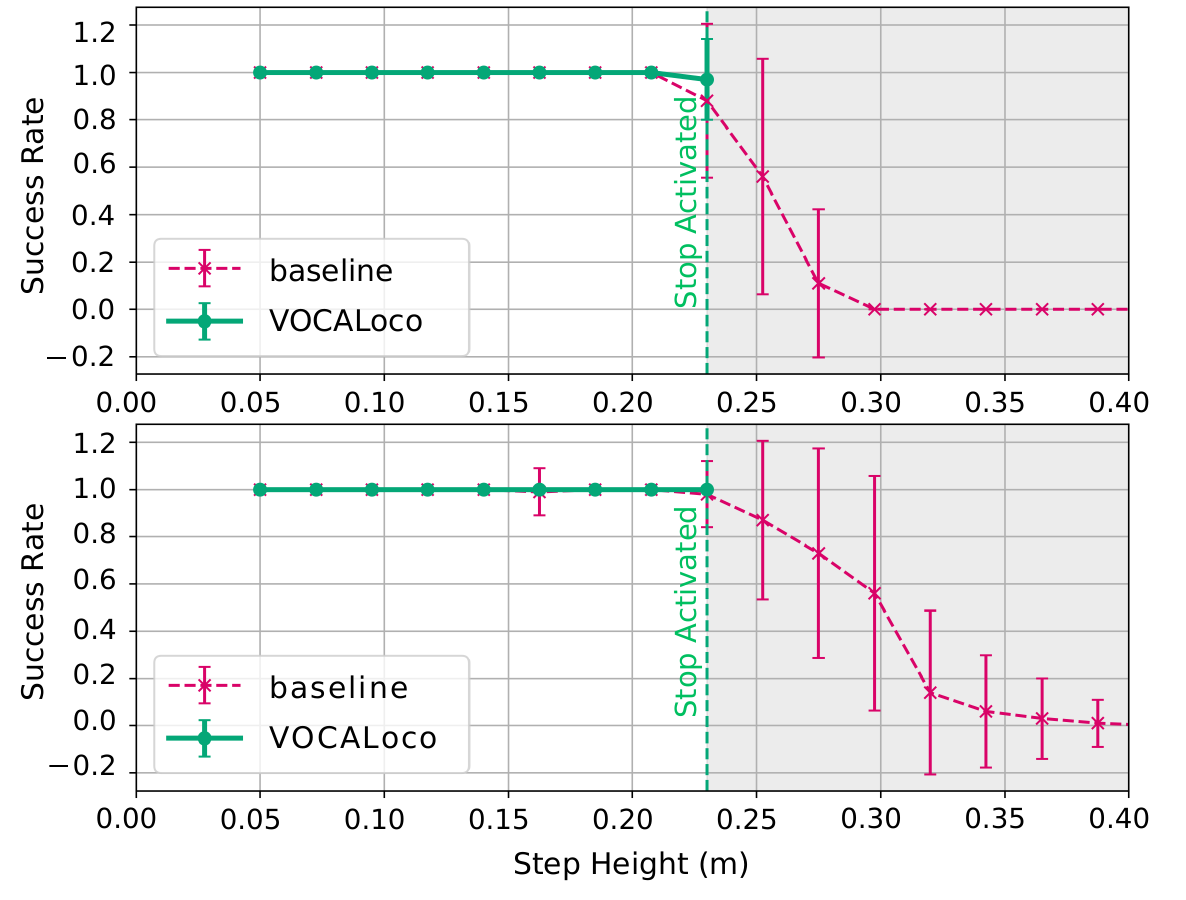}
    \caption{
    \stanley{Success rate of executing policies on terrains from (top) Figure~\hyperref[fig:sim]{\ref*{fig:sim}.5}  and (bottom) Figure~\hyperref[fig:sim]{\ref*{fig:sim}.6} . We perform 100 rollouts per step height and the error bars represent the standard deviation, and plot the success rate across increasing difficulty against the baseline \cite{legged_gym}. 
\textit{Stop Activated} indicates that our viability module has determined the terrain is unviable and prevented the robot from moving. 
    }}
    \label{fig:sim-success-rate}
\end{figure}

\begin{figure}[t]
    \centering
    \includegraphics[width=1\columnwidth, keepaspectratio]{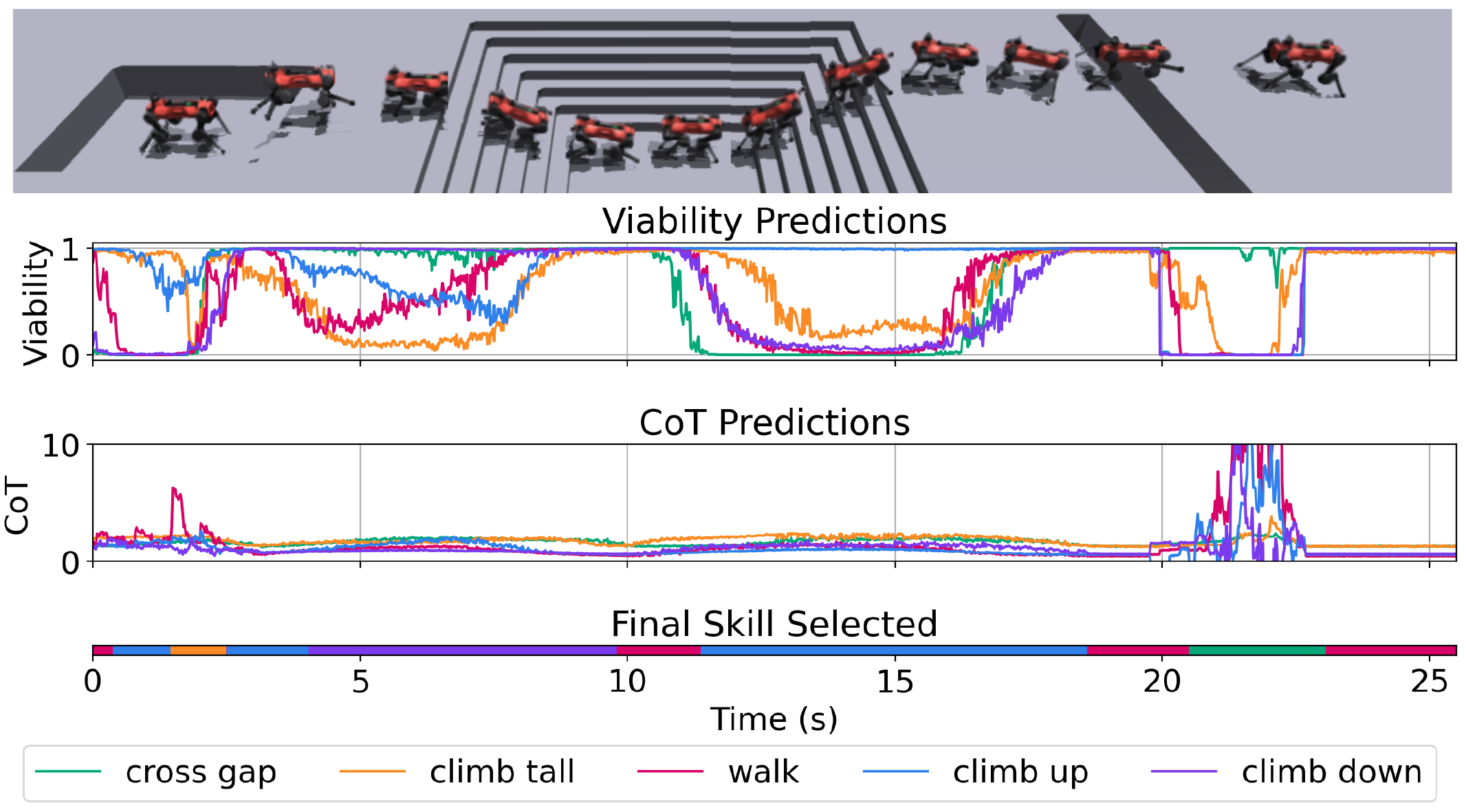}
    \caption{\stanley{Predicted viability, CoT, and final skill to execute on the ANYmal robot on an obstacle course in simulation.}}
    \label{fig:sim-new-skills}
\end{figure}

\begin{figure*}[t!]
    \centering
    \includegraphics[width=\textwidth, keepaspectratio]{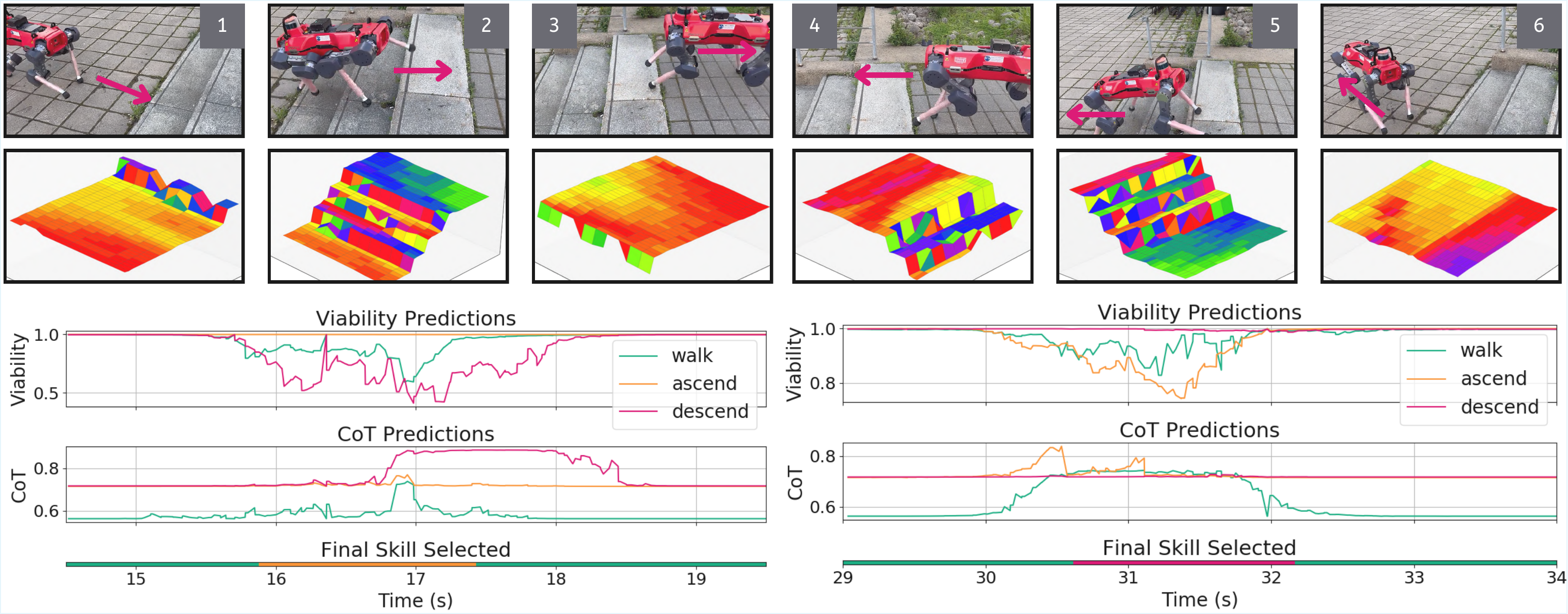}
    \caption{Real-world deployment of ANYmal-D on a staircase with three $12cm$ steps. Snapshots 1-3 show ascent and 4-6 show descent. The rows represent (top) raw frames, (middle) extracted heightfields, and (bottom) viability and CoT predictions. Viability remains above threshold at each policy switch, and CoT curves correctly identify the most efficient policy, confirming smooth, reliable transitions.}
    \label{fig:real-world}
\end{figure*}

\noindent
We want to evaluate the performance of \ours{} against the baseline, a rough terrain ANYmal-D locomotion policy using \cite{legged_gym} in simulation.
Our testing environments are shown in Figures~\hyperref[fig:sim]{\ref*{fig:sim}.5} where the robot spawns on a flat terrain, then ascends stairs, then resumes on flat terrain and Figures~\hyperref[fig:sim]{\ref*{fig:sim}.6} where the robot spawns on a flat terrain, descends stairs, then resumes walking on flat terrain. 
{\em Success} corresponds to reaching the target, depicted by a star in both figures. 


Figure \ref{fig:sim-success-rate} illustrates success rates for scenarios using \cite{legged_gym} as a baseline. 
\textit{Stop Activated} indicates that our viability module has determined the terrain to be unviable, thereby preventing the robot from proceeding.
We observe that \ours{} identifies unviable terrain and halts, whereas the baseline, lacking any untraversability or risk detection, continues tracking the desired velocity and ultimately becomes stuck or crashes on higher-step stairs during both ascent and descent.

\stanley{We note that our low-level policies are more robust on terrains with higher steps than the baseline policy. Looking at the step height of $22.5cm$ for both subplots of Figure~\ref{fig:sim-success-rate}, we either observe fewer crashes or achieve a 100\% success rate at reaching the target. }



\subsection{Modular and Scalable Skill Integration}
\noindent
 \stanley{
To demonstrate that our proposed framework is modular and scalable, we extend our framework to two new skills: climbing tall obstacles and crossing gaps.
 We start off with 3 existing skills, along with their associated viability and CoT networks. Then, as the first step in the extension, we train two new policies, climbing tall obstacles and gap-crossing, with the difference that they take position-based target commands \cite{rudin2022advanced} \cite{anymal-parkour}. Once these skills are trained, we proceed to collect viability and CoT data and train their CNN modules. 
 {\it Without retraining any of the existing CNNs}, we show that direct integration of the new skills and CNNs with the existing framework allows the robot to traverse an obstacle course with a wall and a gap of 0.5 meters (see Figure~\ref{fig:sim-new-skills}). 
}

\subsection{Real World Deployment}

\noindent
We validate \ours{} on real hardware using ANYbotics ANYmal-D \cite{anymal-d}. At all times, a joystick provides linear velocity commands of between $0.5$ and $0.6 m/s$. As in simulation, both the high-level and low-level policies operate at $50$ Hz, and \emph{no parameters are altered} between the simulated and real-robot setups.

Figure~\ref{fig:real-world} shows the output viability and CoT module outputs during deployment on a staircase with $3$ steps all around $12cm$.
Initially, all policies are equally viable, so the policy with a lower CoT is selected, corresponding to the walk policy. As the robot moves toward the staircase, it switches to the ascend policy because the viability module estimates that it is the only viable policy. Once the base of the robot is back on flat terrain, it resumes to the walking policy, as it has the lowest predicted CoT. We note that this switch happens while the hind legs are still on the last step of the stairs. In any of \ours{} runs, however, this does not cause the robot to get stuck, as the walking policy is able to lift its hind legs to clear the last step. Similarly, we can see the descent policy is selected while the robot is descending the stairs, and is using the walking policy on flat terrain. We also do not observe any issues during the transitions between locomotion policies. For more hardware experiment outcomes on other staircases with different heights, please refer to the supplemental video \stanley{available at: \url{https://sites.google.com/view/vocaloco}}.

\section{CONCLUSION}
\noindent

\noindent \stanley{We present a novel approach to switch between terrain-specialized locomotion policies to enhance overall safety and efficiency. Our system combines high-level decision modules, a viability module and a CoT module, with low-level control policies. Each module leverages CNNs to assess the terrain for a fixed horizon. The viability module identifies which policies can safely traverse it, and then, the CoT module selects the most energy-efficient policy. We validated our approach in simulation and successfully deployed it zero-shot on ANYmal-D quadruped. As future work, we plan to extend our framework from discrete skill switching to policy mixing, enabling smoother and more continuous transitions between skills, for example, using a multi-expert composition approach \cite{multi-expert}.}

\bibliographystyle{IEEEtran}
\bibliography{bibtex_files/IEEEabrv,bibtex_files/bibliography}

\end{document}